\documentclass[letterpaper, 10 pt, conference]{ieeeconf}  

\IEEEoverridecommandlockouts                              

\usepackage{amsmath}
\usepackage{graphicx}
\usepackage{multirow}
\usepackage{amssymb}
\usepackage{pifont}
\usepackage{etoolbox}
\usepackage{soul, color, xcolor}
\usepackage{float}

\makeatletter
\patchcmd{\@makecaption}
  {\scshape}
  {}
  {}
  {}
\makeatletter
\patchcmd{\@makecaption}
  {\\}
  {.\ }
  {}
  {}
\makeatother

\title{\LARGE \bf
MTDP: A Modulated Transformer based Diffusion Policy Model
}

\author{Qianhao Wang$^{1,2}$ and Yinqian Sun$^{1}$ and Enmeng Lu$^{1}$ and Qian Zhang$^{1,2,4 *}$ and Yi Zeng$^{1,2,3,4 **}$
\thanks{*Corresponding author. Email: q.zhang@ia.ac.cn}
\thanks{**Corresponding author. Email: yi.zeng@ia.ac.cn}
\thanks{$^{1}$Brain-inspired Cognitive Intelligence Lab, Institute of Automation, Chinese Academy of Sciences, Beijing, China.}
\thanks{$^{2}$School of Artiﬁcial Intelligence, University of Chinese Academy of Sciences,Beijing, China.}
\thanks{$^{3}$Key Laboratory of Brain Cognition and Brain-inspired Intelligence Technology, CAS, Shanghai, China.}
\thanks{$^{4}$Center for Long-term Artiﬁcial Intelligence, Beijing, China.}
}

\begin{document}

\maketitle
\thispagestyle{empty}
\pagestyle{empty}

\begin{abstract}


Recent research on robot manipulation based on Behavior Cloning (BC) has made significant progress. By combining diffusion models with BC, diffusion policiy has been proposed, enabling robots to quickly learn manipulation tasks with high success rates. However, integrating diffusion policy with high-capacity Transformer presents challenges, traditional Transformer architectures struggle to effectively integrate guiding conditions, resulting in poor performance in manipulation tasks when using Transformer-based models. In this paper, we investigate key architectural designs of Transformers and improve the traditional Transformer architecture by proposing the Modulated Transformer Diffusion Policy (MTDP) model  for diffusion policy. The core of this model is the Modulated Attention module we proposed, which more effectively integrates the guiding conditions with the main input, improving the generative model's output quality and, consequently, increasing the robot's task success rate. In six experimental tasks, MTDP outperformed existing Transformer model architectures, particularly in the Toolhang experiment, where the success rate increased by 12\%. To verify the generality of Modulated Attention, we applied it to the UNet architecture to construct Modulated UNet Diffusion Policy model (MUDP), which also achieved higher success rates than existing UNet architectures across all six experiments. The Diffusion Policy uses Denoising Diffusion Probabilistic Models (DDPM) as the diffusion model. Building on this, we also explored Denoising Diffusion Implicit Models (DDIM) as the diffusion model, constructing the MTDP-I and MUDP-I model, which nearly doubled the generation speed while maintaining performance.

\end{abstract}

\section{INTRODUCTION}
Behavior Cloning \cite{pomerleau1988alvinn} is one of the methods of imitation learning \cite{ng2000algorithms} and serves as a simple and intuitive approach for robot control by directly learning the mapping from states to actions. This method does not require additional complex structures \cite{merel2017learning,stadie2017third} or the challenging architectures associated with reinforcement learning \cite{mnih2013playing,lillicrap2015continuous,schulman2017proximal}. However, it suffers from poor generalization and the issue of learning overly simplistic behavior patterns from expert demonstration data \cite{ross2011reduction,hussein2017imitation}. To address these issues, a novel visual motion control policy, Diffusion Policy (DP) \cite{chi2023diffusion}, has been recently proposed. This policy combines diffusion models with BC methods to generate executable actions through iterative denoising guided by visual inputs. Diffusion models are capable of naturally capturing multiple behavior patterns present in the demonstration data, thereby producing diverse action sequences. This diversity enables the policy to flexibly select the most suitable actions for varying environments and tasks, thereby enhancing generalization capabilities. The proposed method constructs models based on both Transformer \cite{vaswani2017attention} and UNet \cite{ronneberger2015u} architectures. However, model based on the Transformer architecture has shown poor performance in tasks due to their inability to adequately integrate main inputs with guiding conditions. By studying and modifying the traditional Transformer architecture, we have proposed a Transformer-based model, MTDP, tailored for the diffusion policy. Validation across six experimental tasks demonstrates that this architecture effectively integrates with diffusion models, improving the success rate of robot manipulation tasks.

In recent years, the Transformer has been widely applied in the fields of natural language processing \cite{radford2019language,alaparthi2021bert} and computer vision \cite{alexey2020image,radford2021learning} due to its advantages of high capacity and scalability, achieving significant progress such as Large Language Models (LLMs) \cite{achiam2023gpt,touvron2023llama} and Vision-Language Models (VLMs) \cite{oquab2023dinov2,zhai2023sigmoid,kirillov2023segment}. Inspired by this, the Transformer has gradually been adopted in the robotics field \cite{zitkovich2023rt,vuong2023open,kim2024openvla,brohan2022rt}. The traditional Transformer architecture consists of an encoder-decoder structure, where the encoder is primarily composed of self-attention modules and feedforward neural networks, and the decoder mainly includes self-attention modules, cross-attention modules, and feedforward neural networks.

In the DP \cite{chi2023diffusion}, guiding conditional features are initially fed into the encoder and subsequently into the decoder to guide the generation of action trajectories. However, these guiding conditions are only fused with the input through cross-attention modules, which is insufficient and does not fully utilize them. Drawing inspiration from the Diffusion Transformer (DIT) \cite{peebles2023scalable} structure, we integrated it with the Transformer to propose the Modulated Attention Module (the core of the MTDP model) for replacing the original structure of the decoder. This attention module can fully leverage the guiding conditions, enhancing the quality of generation and thereby increasing the success rate of task execution. To verify its versatility, we also applied it to the UNet architecture and conducted experiments, achieving superior performance to the original UNet architecture in DP across all experimental tasks.

Additionally, in the diffusion policy, the diffusion model typically uses DDPM model \cite{ho2020denoising}. Since DDIM \cite{song2020denoising} can ensure generation quality while reducing the generation time compared to DDPM, we also explored the effects of using DDIM as the diffusion model of the diffusion policy.

\begin{figure*}[htbp]
	\centering
	\includegraphics[width=0.8\textwidth]{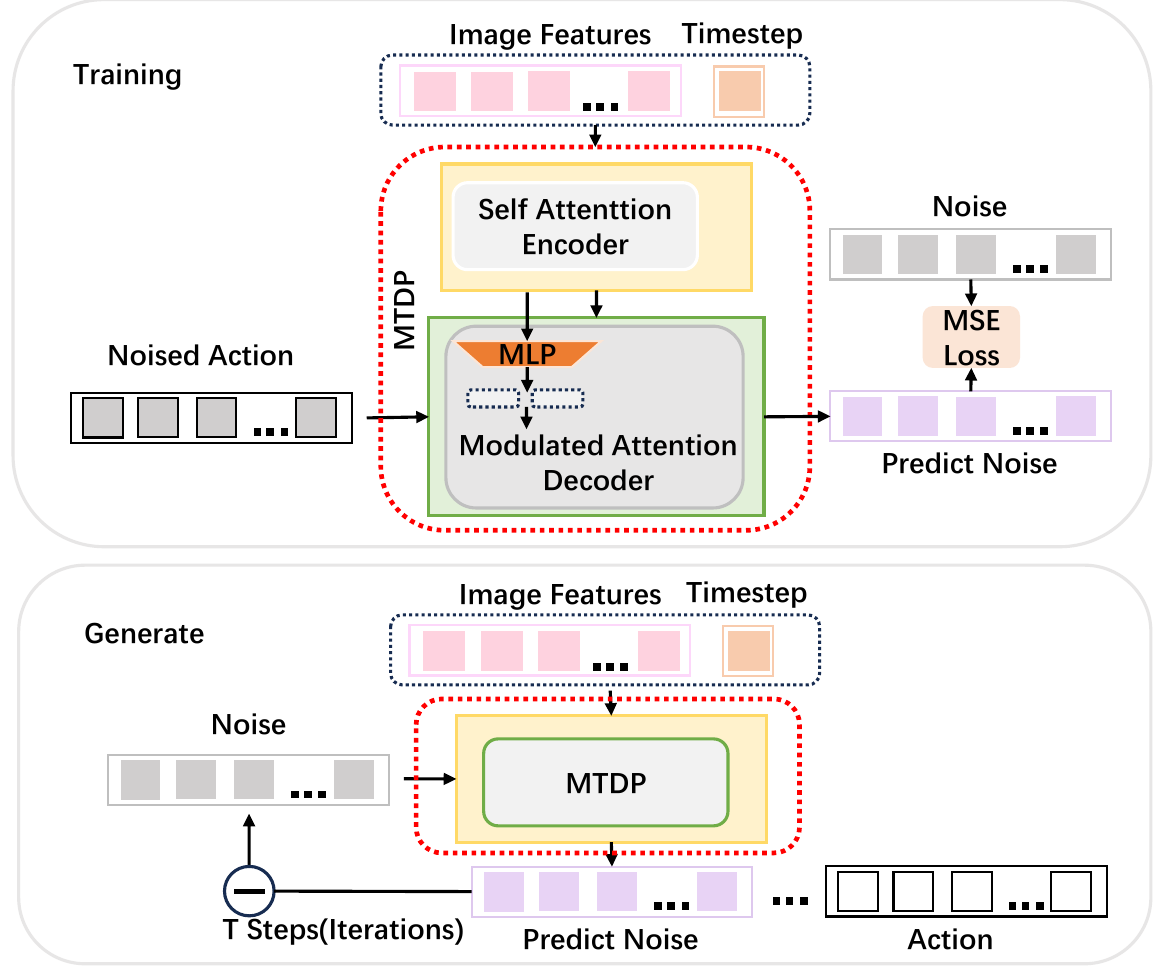} %
	\caption{The architecture of MTDP. Both the timestep and image features serve as guiding conditions and are input into the encoder. During the training phase, guiding conditions combined with noisy action features and fed into the decoder; while during the generation phase, combined with noise. Through iterative denoising, executable actions are obtained. }
	\label{fig1}
\end{figure*} 

In summary, the main contributions of this paper are as follows:

\begin{itemize}

\item We propose MTDP, a Transformer-based diffusion policy model, which improving the traditional Transformer decoder structure by introduceing the Modulated Attention module. This model effectively integrating guiding conditions with the main input, fully leveraging the guiding conditions to perform tasks. 

\item
We test the proposed MTDP model across six experimental tasks. The success rate of our model exceeded that of the existing Transformer models in almost all experiments, particularly achieving a 12\% improvement in the Toolhang task. This demonstrates the effectiveness of our proposed architecture.

\item

We substituted the DDPM with DDIM in our diffusion model to develop the MTDP-I model, which was also evaluated across six experimental tasks. Compared to the MTDP model, the MTDP-I model showed a slight decrease in success rate, while the generation timestep (sampling timestep) was reduced by 40\%. This demonstrates that the MTDP-I model significantly reduces generation time while maintaining experimental effectiveness.

\end{itemize}

\begin{figure*}[htbp]
	\centering
	\includegraphics[width=0.9\textwidth]{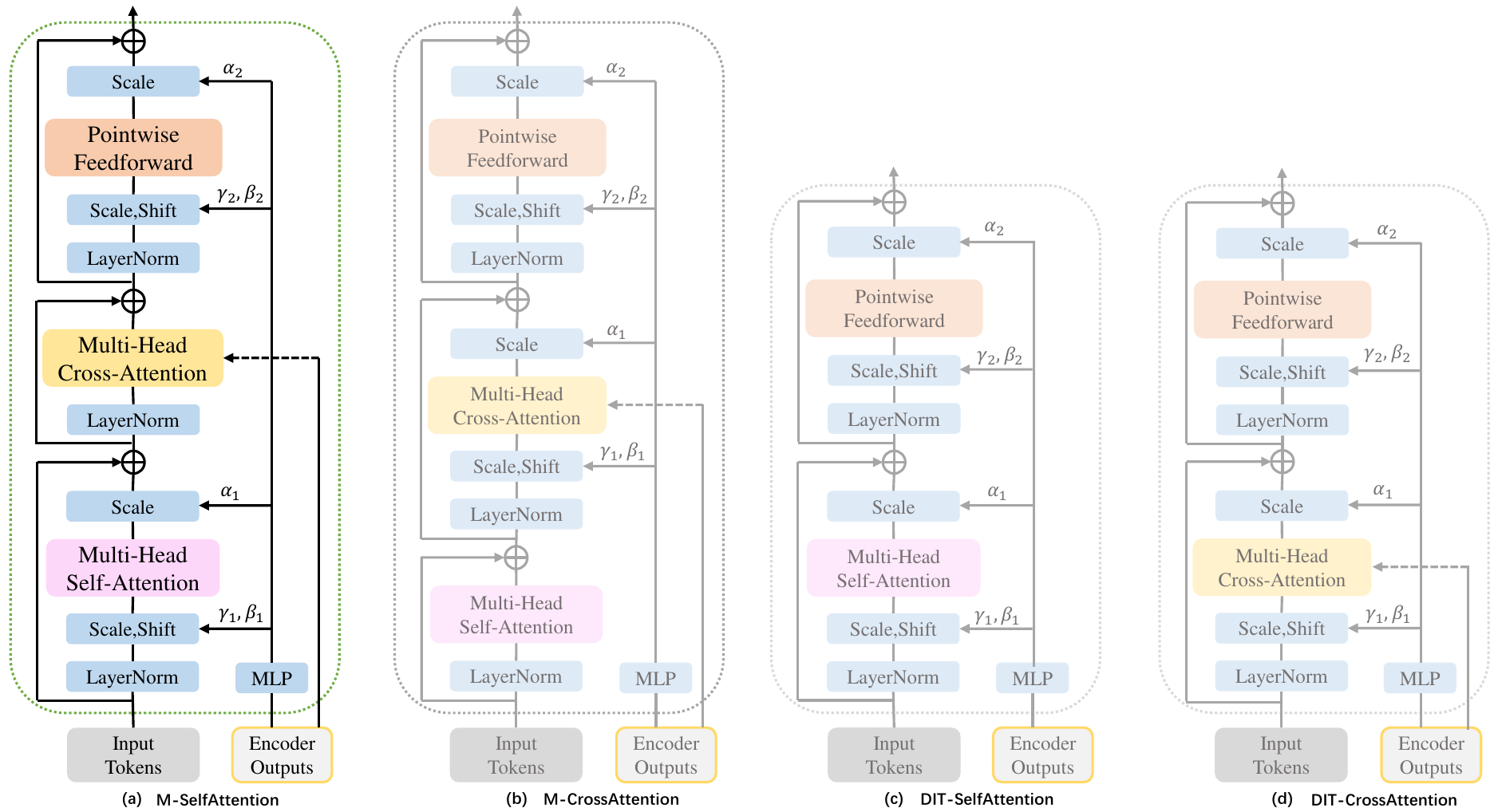} %
	\caption{We experimented with four different structures for Modulated Attention. (a) M-SelfAttention represents a model with both modulated self-attention and cross-attention; (b) M-CrossAttention represents a model with both self-attention and modulated cross-attention; (c) DIT-SelfAttention is DIT, represents a model with only modulated self-attention; (d) DIT-CrossAttention represents a model with only modulated cross-attention. We found that (a) performed the best. }
	\label{fig2}
\end{figure*}

\section{RELATED WORK}

Diffusion Models: Diffusion models operate on the fundamental principle of simulating the gradual degradation (by adding noise) and subsequent restoration (through denoising) of data. Due to their realistic and diverse generation capabilities, diffusion models have undergone rapid advancements recently. This progress has led to the development of various notable models, including DDPM \cite{ho2020denoising} and DDIM \cite{song2020denoising}. Peebles et al. \cite{peebles2023scalable} deviated from the traditional UNet-based diffusion models by proposing a Transformer-based diffusion model, DIT, which achieved improved generation results. Unlike the DIT, which retains the encoder architecture, our model maintains an encoder-decoder structure and modifies the decoder to better integrate the guiding conditions.

Diffusion Policy: DP \cite{chi2023diffusion} involves applying diffusion models to robot control, generating executable action trajectories for robots. This method can learn multiple behavior patterns from demonstration data, offering better generalization and achieving widespread adoption in the robotics field. The STMDP model \cite{wang2024brain} applies diffusion policy to spiking neural networks, constructing a robot control model based on the Spiking Transformer. The OCTO model \cite{team2024octo} develops a novel Transformer-based universal robot policy model and incorporates a lightweight action head to implement diffusion policy, effectively converting model outputs into actions. The Robotics Diffusion Transformer (RDT) \cite{liu2024rdt} introduces a physically interpretable unified action space that standardizes action representations across various robots while preserving the physical meanings of the original actions. In terms of model architecture, RDT builds a diffusion foundational model based on the Transformer and incorporating a branch structure from DIT. Compared to OCTO, which employs a lightweight action head to realize diffusion policy, Hou et al. \cite{hou2024diffusion} developed a large-scale diffusion policy model based on DIT for universal robot policy learning. Our work focuses on exploring the efficient structure of Transformers suitable for the diffusion policy.


\begin{figure*}[htbp]
	\centering
	\includegraphics[width=0.9\textwidth]{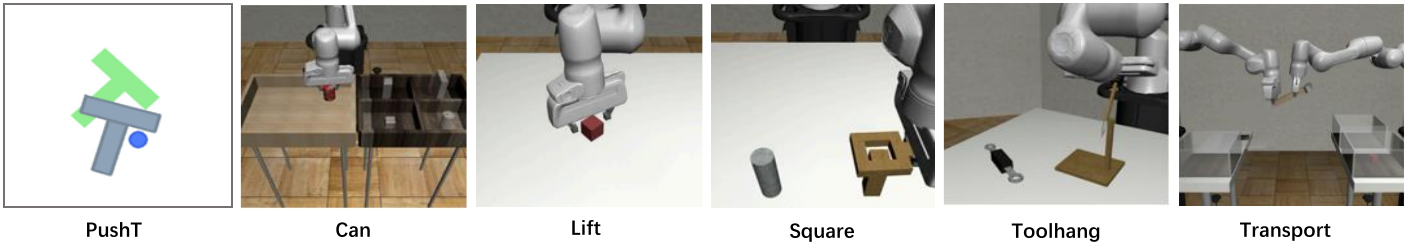} %
	\caption{Tasks used for the experiment. }
	\label{fig3}
\end{figure*}

\section{METHOD}

In this section, we provide a detailed analysis of the MTDP and MUDP models, which are diffusion policy models based on Transformer and UNet architectures, respectively. Both MTDP and MUDP can enhance the success rate of robots performing manipulation tasks. Additionally, we present the Modulated Attention module, a core component fundamental to both the MTDP and MUDP models.

\subsection{ The Architecture Of MTDP And MUDP}
We propose MTDP, a Transformer-based architecture for robot diffusion policy. The specific structure of MTDP is shown in Figure \ref{fig1}. We feed the image and timestep features, along with the noisy action, into MTDP, which outputs the predicted noise. MTDP consists of an encoder with self-attention and a decoder with Modulated Attention. The encoder remains unchanged, while we replace the multi-head self-attention, multi-head cross-attention, and feedforward neural network structure in the decoder with Modulated Attention. To verify the universality of Modulated Attention, we construct the UNet-based architecture MUDP and modify both the upsampling and downsampling modules of UNet. Each of these modules contains two conditional convolution blocks, with one conditional convolution block replaced by a Modulated Attention block, while keeping the rest of the structure unchanged.

\subsection{The Architecture of Modulated Attention}

We propose the Modulated Attention with the objective to make fuller use of guiding conditions, enhance the generative effects of diffusion policy, and subsequently improve the success rate of robots performing manipulation tasks. Figure \ref{fig2} displays the specific structure of Modulated Attention, and in this section, we will detail the various components of Modulated Attention. The input to the Modulated Attention module comes from the decoder input as well as the encoder's output (guiding conditions). We have experimented with four different Modulated Attention structures that integrate guiding conditions in diverse ways.

\begin{itemize}

\item Modulated Self-Attention (M-SelfAttention): The encoder outputs are mapped through an MLP layer to generate modulation parameters, which then modulate both the self-attention layers and feedforward neural network layers. The cross-attention layers receive outputs from the encoder and direct integrates the guiding conditions. In this structure, both the self-attention layers and cross-attention layers integrate the guiding conditions, which makes fuller use of the guiding conditions.

\item Modulated Cross-Attention (M-CrossAttension): This structure utilizes modulated parameters to modulate the cross-attention layers and the feedforward neural network layers. Additionally, the cross-attention layers receive direct guidance from the guiding conditions. But the self-attention layers remain unchanged and does not integrate guiding conditions.

\item DIT for Self-Attention (DIT-SelfAttention): This structure follows the DIT architecture; however, the distinction lies in this module serving as the core component of the Transformer decoder and used for diffusion policy. Unlike M-SelfAttention, this structure does not have the cross-attention layer.

\item DIT for Cross-Attention (DIT-CrossAttension): This structure is similar to the DIT, except that the modulation module shifts from self-attention to cross-attention. In this structure, only the cross-attention integrates the guiding conditions. Unlike M-CrossAttension, this structure does not include a self-attention layer.

\end{itemize}

\subsection{The Overall Training Process}

The diffusion policy process is divided into training phase and generation phase. The training phase aims to train a noise prediction network, while the generation phase utilizes the trained noise prediction network to predict noise, iteratively converting random noise into robot-executable actions. We proposed MTDP and MUDP models serve as the noise prediction network $\epsilon_\theta(x_t,c_{obs},t)$, where $x_t\in \mathcal{N}(0,I)$  represents the noise vector at timestep $t$, $c_{obs}$ denotes image observations, and $t$ represents the timestep. During the training stage, noise $x_t$ is added to the action $a$ to become $\overline{a}$, and the noise prediction network $\epsilon_\theta(\overline{a},c_{obs},t)$ predicts the noise $\overline{x}_t$, with the timestep $t$ randomly selected. The network weights are updated by computing the mean squared error (MSE) loss between $x_t$ and $\overline{x}_t$. In the generation process, the timestep $T$ is fixed, and random Gaussian noise $x_T$ is introduced into the noise prediction network; guided by the image conditions, $T$  iterations of denoising are performed to generate executable actions. 

DDPM \cite{ho2020denoising} transform the task from predicting noisy images to predicting the noise itself, thereby reducing prediction difficulty and enhancing generation performance. DDIM \cite{song2020denoising} manually construct the posterior distribution $q(x_{t-1}\mid x_t,x_0)$, ensuring that for $t>=2$,$q(x_t\mid x_0) = \mathcal{N}(x_t;\sqrt{\overline{\alpha}_t}x_0,(1-\overline{\alpha}_t)I)$, ensuring that the marginal distribution under this distribution remains Gaussian, while the forward process no longer adheres to a Markov process. This characteristic allows the sampling process to operate without the need for a large number of sampling timestep, enabling the generation of high-quality images with fewer sampling timestep. In the diffusion policy, the primary diffusion model currently used is DDPM. We have conducted detailed research on DDIM as the diffusion model within diffusion policy. By performing an ablation study on its timestep, we have nearly doubled the speed of the generation process while maintaining its effectiveness. The generative processes based on DDPM and DDIM are described in Equation \eqref{eq1} and Equation \eqref{eq2}, respectively.

\begin{equation}
\label{eq1}
	x_{t-1} = \sqrt{\alpha_t} \left( x_t - \frac{1 - \alpha_t}{\sqrt{1 - \overline{\alpha}_t}} \epsilon_\theta(x_t,c_{obs},t) \right) + \mathcal{N}(0,\sigma^2I)
\end{equation}

\begin{multline}
\label{eq2}
	x_{t-1} = \sqrt{\alpha_{t-1}} \left( \frac{1}{\sqrt{\alpha_{t}}} x_t - \frac{1 - \alpha_t}{\sqrt{\alpha_{t}}} \epsilon_\theta(x_t, c_{obs}, t) \right) +\\ \sqrt{1 - \alpha_{t-1}} \epsilon_\theta(x_t, c_{obs}, t)
\end{multline}
where, $\alpha_t,\overline{\alpha}_t,\sigma$ are all noise scheduling coefficients.

\begin{table*}[htbp]
\begin{center}
\caption{Comparison with baseline methods, including the DP-Transformer and DP-DIT. MTDP outperforms both DP-Transformer and DP-DIT across nearly all experimental tasks. The MTDP-I improves performance while concurrently reducing the number of timesteps involved in the generation process. The asterisk (${\ast}$) indicates results obtained from re-experiments conducted using the methods described in the original paper.}
\label{table1}
\begin{tabular}{c|c|c|c|c|c|c|c|c}
\hline
\multirow{2}{*}{Model} &\multirow{2}{*}{MC-Attention}&\multirow{2}{*}{Timestep}&\multicolumn{6}{c}{Tasks}\\
\cline{4-9}

 &&& PushT & Can & lift & Square & Toolhang & Transport \\
\hline

DP-Transformer\cite{chi2023diffusion}$^{\ast}$ &\ding{55}&100&0.70&\textbf{0.98} &1.0 &0.93	&0.60	&0.85\\ 
DP-DIT &\ding{55}&100& 0.72 &0.97&1.0 &\textbf{0.95} &0.70	&0.91\\ 
\hline
MTDP(ours)     & \checkmark &100& \textbf{0.74}(+4\%) &0.97(-1\%)&1.0 &0.93	&\textbf{0.72}(+12\%)	&\textbf{0.92}(+7\%)   \\ 
MTDP-I(ours)  & \checkmark & \textbf{60}& 0.73(+3\%) & \textbf{0.98} &1.0 & \textbf{0.95}(+2\%)	&0.68(+8\%)&0.88(+3\%)\\ 
\hline
\end{tabular}
\end{center}
\end{table*}

\begin{table*}[htbp]
\begin{center}
\caption{To demonstrate the versatility of Modulated Attention, we apply our method to the UNet architecture and compare it with the DP-UNet. MUDP outperforms DP-UNet across all experimental tasks, demonstrating the versatility of Modulated Attention. MUDP-I also enhances performance while simultaneously reducing the timesteps required in the generation process. The asterisk (*) indicates results obtained from re-experiments conducted using the methods described in the original paper. }
\label{table2}
\begin{tabular}{c|c|c|c|c|c|c|c|c}
\hline
\multirow{2}{*}{Model} &\multirow{2}{*}{MC-Attention}&\multirow{2}{*}{Timestep}&\multicolumn{6}{c}{Tasks}\\
\cline{4-9}

 &&& PushT & Can & lift & Square & Toolhang & Transport \\
\hline

DP-UNet\cite{chi2023diffusion}\^* &\ding{55}&100&0.84&0.98 &1.0 &0.93	&0.80	&0.89\\ 
\hline
MUDP(ours)     & \checkmark &100& 0.84 &0.98&1.0 &\textbf{0.94}(+1\%)	&\textbf{0.81}(+1\%)	&\textbf{0.9}(+1\%)   \\ 
MUDP-I(ours)  & \checkmark &\textbf{60}& \textbf{0.85}(+1\%) & \textbf{0.99}(+1\%) &1.0 & \textbf{0.94}(+1\%)	&\textbf{0.81}(+1\%)&0.89\\ 
\hline
\end{tabular}
\end{center}
\end{table*}

\section{EXPERIMENT}

We compared our proposed method with two baseline approaches, one is diffusion policy based on the Transformer architecture \cite{chi2023diffusion}, another is diffusion policy based on DIT \cite{peebles2023scalable} which we constructed, across six robotic manipulation tasks to evaluate the effectiveness of our MTDP architecture and the efficacy of Modulated Attention. Additionally, to verify the universality of Modulated Attention, we constructed the MUDP architecture and conducted experiments, demonstrating that Modulated Attention is not only applicable to the Transformer architecture but also effective on the UNet architecture. To determine the components of Modulated Attention, we constructed four different structures and tested them on PushT, Square and Toolhang experimental tasks. Finally, to increase the speed of the generation process, we also investigated the experimental performance of DDIM as the diffusion model within the diffusion policy.

\subsection{Experimental Setup}

We conducted experiments on six robot manipulation tasks \cite{chi2023diffusion}, including PushT, Can, Lift, Square, Toolhang, and Transport. These six tasks involve robotic manipulation in both 2D planes and 3D spaces, as shown in Figure \ref{fig3}. Diverse experimental tasks make the model evaluation more convincing.

\begin{itemize}

\item PushT: Use a circular end effector to push a T-shaped block to a blue T-shaped marker on the table.
\item Can: Grasp a can of soda and place it in a slot next to the table.
\item Lift: Pick up a cube from the table.
\item Square: Grasp an object with a square hole and install it onto a square pillar.
\item Toolhang: Assemble a rack from a set of tools on the table and hang the remaining tools on the rack.
\item Transport: Two robotic arms collaborate to pass an object.

\end{itemize}

When using DDPM as the diffusion model, the experimental parameters were set to match those in DP \cite{chi2023diffusion}, with the number of timestep $T$ set to 100. When using DDIM as the diffusion model, timestep $T$ was set to 60, increase the speed of generate process. We used three random seeds and averaged the results for testing.

\begin{figure}[htbp]
	\centering
	\includegraphics[width=0.4\textwidth]{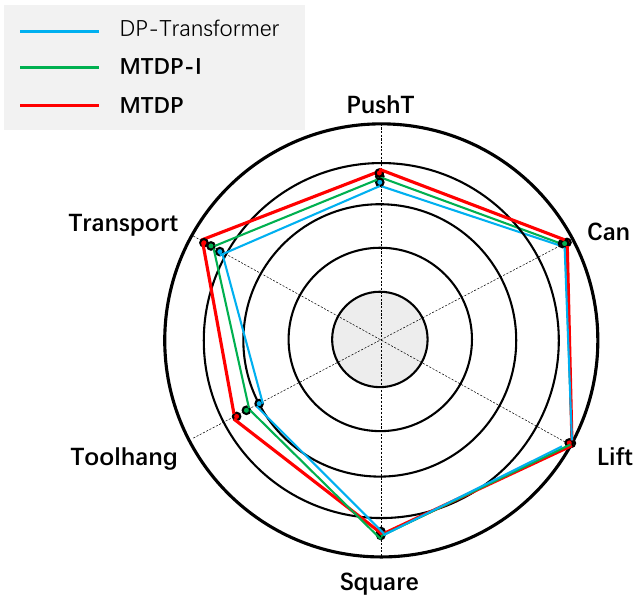} %
	\caption{Comparison of task execution performance between existing Transformer-based diffusion policy models and our proposed model. }
	\label{fig4}
\end{figure}

\subsection{Experiment Result}

In this section, we evaluate the performance of MTDP and MUDP on robotic manipulation tasks, as well as the model's performance when replacing the diffusion model from DDPM to DDIM.

For the MTDP model, we use the diffusion policy based on the Transformer architecture (DP-Transformer) \cite{chi2023diffusion} as a baseline method for comparison. To comprehensively measure the advantages of our model, we also construct and implement a diffusion policy based on the DIT \cite{peebles2023scalable} architecture (DP-DIT), which is used as another baseline for comparison. DP-DIT replaces the Transformer architecture with the DIT architecture, including both the encoder and decoder, rather than just replacing the decoder part, while keeping the rest unchanged. The experimental results are shown in Table \ref{table1}, where MTDP outperforms both the DP-Transformer and DP-DIT methods in nearly all experimental tasks, particularly in the Toolhang experiment, with an improvement of 12\%, and an overall average increase of 4\%. For the MUDP model, we compare it against the DP based on the UNet architecture (DP-UNet) \cite{chi2023diffusion}. The experimental results, as shown in Table \ref{table2}, indicate that MUDP is equal to or exceeds the DP-UNet method in all experimental tasks.

From Figure \ref{fig4}, it becomes clear that our Transformer-based diffusion policy architecture significantly surpasses the performance of existing Transformer-based diffusion policy. This highlights the advantages of our Modulated Attention module, which more effectively integrates guiding conditions to steer the generation process and improve the output quality. Although the enhancements provided by Modulated Attention are less pronounced when utilized with the UNet architecture compared to its application in the Transformer architecture, this also demonstrates the module's broad applicability. We have developed the diffusion policy models MTDP-I and MUDP-I, utilizing DDIM as the diffusion model, whereas previous models employed DDPM. Although the performance enhancements in MTDP-I are not as marked as those in the MTDP model, it still surpasses the existing Transformer-based diffusion policy. The MUDP-I model performs nearly on par with the MUDP model and surpasses the diffusion policy based on the UNet architecture. Furthermore, the forward generation process of MTDP-I and MUDP-I requires only 60 timesteps, which reduces 40\% compared to 100 timesteps utilized by DDPM as a diffusion model within the diffusion policy. This highlights the advantages of DDIM as a diffusion model, offering comparable performance while requiring less time for inference.

\subsection{Ablation Study}

In this section, we perform an ablation study on the key structural designs of Modulated Attention, and an ablation study on the timestep $T$ of DDIM.



\begin{itemize}
\item Ablation Study on four configurations of Modulated Attention
\end{itemize}

Through an in-depth study of Transformer and DIT architectures, we constructed four distinct configurations of the decoder core module Modulated Attention, as illustrated in Figure \ref{fig2}. Each configuration employs a different approach to handling guiding conditions. We evaluated these structures on three experimental tasks, PushT, Square and Toolhang, using the Transformer architecture, with results presented in Table \ref{table3}. The DIT-SelfAttention structure, lacking a cross-attention layer, performed poorly on the Pusht and Toolhang tasks, underscoring the indispensability of the cross-attention. The DIT-CrossAttention structure, lacking a self-attention layer, demonstrated slightly inferior overall performance across both tasks compared to the M-SelfAttention methods. While DIT-CrossAttention achieved a 1\% higher success rate than M-SelfAttention in the Toolhang task, we observed instances of extremely low success rates in its validation results, accompanied by high variance. Given this inconsistency, we ultimately chose not to adopt this structure. Additionally, the M-CrossAttention showed weaker performance in the ToolHang task. We think that the cross-attention layer in the M-CrossAttention structure, already directly guided by conditions from the encoder, might result in some redundancy when further modulated by guiding parameters.

\begin{table}[h]
\begin{center}
\caption{Ablation study of the Modulated Attention structure, constructing four different configurations that integrate guiding conditions in various ways. The configuration that best utilizes the guiding conditions is selected as the Modulated Attention structure.}
\label{table3}
\begin{tabular}{c|c|c|c|c}
\hline
\multicolumn{2}{c}{\multirow{2}{*}{Architecture} }&\multicolumn{2}{|c}{Tasks}\\
\cline{3-5}

\multicolumn{2}{c|}{}& PushT & Square&Toolhang\\
\hline
\multirow{4}{*}{Transformer\_based}   &DIT-SelfAttention&0.62 & \textbf{0.93}&0.61\\ 
&DIT-CrossAttention&  0.71 & 0.92&\textbf{0.73}\\ 
&M-CrossAttention&  0.73 &\textbf{0.93}&0.63\\
\cline{2-5}
&M-SelfAttention&  \textbf{0.74}&\textbf{0.93}&0.72\\ 

\hline
\end{tabular}
\end{center}
\end{table}


 However, in the M-SelfAttention configuration, the self-attention layer, initially not guided by the encoder's conditions, now integrates guiding parameters with the conditions, while the cross-attention layer also receives inputs from the guiding conditions. This enables the decoder to fully leverage the encoder's guiding conditions, thereby enhancing generation performance. Based on these findings, we ultimately selected the M-SelfAttention configuration as the structure for Modulated Attention.

\begin{figure}[htbp]
	\centering
	\includegraphics[width=0.3\textwidth]{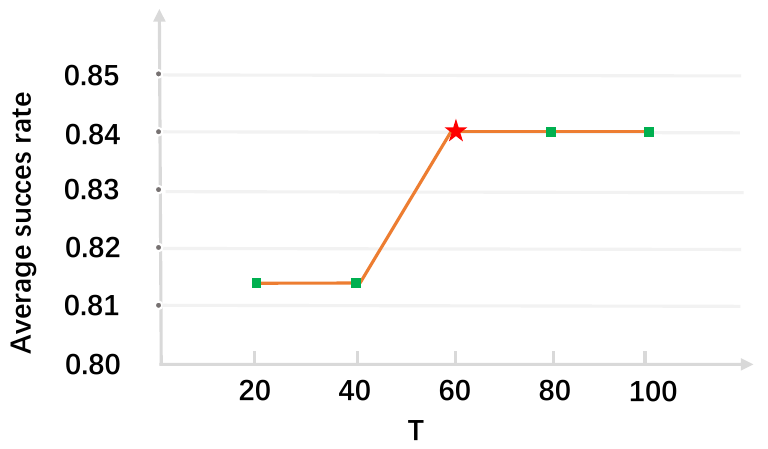} %
	\caption{Average success rates of the MUDP-I model on the PushT and Square tasks at different sampling timesteps. }
	\label{fig5}
\end{figure}

\begin{itemize}
\item Ablation Study on timestep of DDIM
\end{itemize}

DDIM \cite{song2020denoising} can accelerate the generation process, and we investigated the impact of different sampling timestep on experimental results. We set the timestep $T$ at 20, 40, 60, 80, and 100, and conducted tests on two experimental tasks: Pusht and Square. The experimental results are shown in the figure \ref{fig5}. As the number of sampling timestep decreases, the average success rate initially remains unchanged. When reduced below 60, the success rate starts to decline. To balance the experimental performance and the sampling timestep, we ultimately chose $T=60$.
 

\section{CONCLUSIONS}

In this paper, we introduce a Transformer-based diffusion policy model, MTDP, along with the Modulated Attention module. Modulated Attention enables the MTDP model to make fuller use of guiding conditions, thereby enhancing the success rate of task execution. Extensive experiments have demonstrated the effectiveness of our model. Based on MTDP, we proposed MTDP-I, which achieves comparable results but significantly reduces the generation timestep. Additionally, Our MTDP model is scalable, and future work will involve extending it to construct a new Visaion-Language-Action(VLA) model architecture. Given that the Modulated Attention efficiently utilizes guiding conditions, we hope to apply this attention mechanism to other domains, such as imaging.

\subsection{Data Avaliable}
The model of this research is one of the core and part of BrainCog Embot \cite{BrainCogEmbot}. BrainCog Embot is an Embodied AI platform under the Brain-inspired Cognitive Intelligence Engine (BrainCog) framework, which is an open-source Brain-inspired AI platform based on Spiking Neural Network.


\newpage
\section*{ACKNOWLEDGMENT}
This work is supported by the Strategic Priority Research Program of the Chinese Academy of Sciences (Grant No. XDB1010302), the funding from Institute of Automation, Chinese Academy of Sciences (Grant No. E411230101), National Natural Science Foundation of China (Grant No. 62406325), Postdoctoral Fellowship Program of CPSF (Grant No. GZC20232994).


\bibliographystyle{IEEEtran}
\bibliography{IEEEexample} 

\end{document}